\documentclass[10pt,twocolumn,letterpaper]{article}

\usepackage{wacv}
\usepackage{times}
\usepackage{epsfig}
\usepackage{graphicx}
\usepackage{amsmath}
\usepackage{amssymb}

\usepackage{color}
\usepackage{multirow}
\usepackage{enumitem}
\usepackage{subfig}
\usepackage{makecell}

\newcommand{\norm}[1]{\left\lVert#1\right\rVert}

\def\wacvPaperID{334} 

\wacvfinalcopy 

\ifwacvfinal
\usepackage[breaklinks=true,bookmarks=false]{hyperref}
\else
\usepackage[pagebackref=true,breaklinks=true,colorlinks,bookmarks=false]{hyperref}
\fi

\begin{document}

\title{Recovering Trajectories of Unmarked Joints in 3D Human Actions \\ Using Latent Space Optimization}

\author{Suhas Lohit\\
Mitsubishi Electric Research Laboratories\\
Cambridge, MA\\
\and
Rushil Anirudh\\
Lawrence Livermore National Laboratories\\
Livermore, CA\\
\and
Pavan Turaga \thanks{PT partly supported by NSF grant 1452163}\\
Arizona State University, Tempe AZ \\
}

\maketitle
\thispagestyle{empty}

\begin{abstract}
   Motion capture (mocap) and time-of-flight based sensing of human actions are becoming increasingly popular modalities to perform robust activity analysis. Applications range from action recognition to quantifying movement quality for health applications. While marker-less motion capture has made great progress, in critical applications such as healthcare, marker-based systems, especially {\em active markers}, are still considered gold-standard.  However, there are several practical challenges in both modalities such as visibility, tracking errors, and simply the need to keep marker setup convenient wherein movements are recorded with a reduced marker-set. This implies that certain joint locations will not even be marked-up, making downstream analysis of full body movement challenging. To address this gap, we first pose the problem of reconstructing the unmarked joint data as an ill-posed linear inverse problem. We recover missing joints for a given action by projecting it onto the manifold of human actions, this is achieved by optimizing the latent space representation of a deep autoencoder. Experiments on both mocap and Kinect datasets clearly demonstrate that the proposed method performs very well in recovering semantics of the actions and dynamics of missing joints. We will release all the code and models publicly.
\end{abstract}

\section{Introduction}

With the proliferation of low-cost sensing devices, the use of action and gesture recognition has created new capabilities in health tracking, home-based rehabilitation etc. In many of these applications, the essential task at hand is inferring high-level semantic quantities such as the health of the patient, quality of movement, progression of therapy protocols and intended gestures. In these situations, marker-based motion capture, marker-less devices such as Kinect, and more recently wearables like IMU motion sensors, are considered critical to obtain accurate health tracking information. Marker-based mocap systems, especially with active marker sets, which include uniquely colored LEDs per joint, are more accurate. However, they still suffer from self-occlusion and portability problems due to the need to carefully markup many joints.  Also, reducing occlusions would require an expensive setup with a large number of cameras which is not desirable. Devices such as the Kinect have not fully realized their potential in this application space due to similar reasons such as occlusion and tracking errors.  
These errors tend to have a large impact on assessment of human movement quality in health applications \cite{webster2014systematic}. Further, specific conditions such as 
postural deformities may result failure in tracking \cite{regazzoni2014rgb} and in the posture not even being detected \cite{KinectTherapy}. Thus, there is a growing need to use reduced marker-sets, while maintaining the accuracy of tracking afforded by marker-based motion capture systems.

In this paper, we present a framing of this problem from the perspective of inverse problems, deep priors and dynamical systems theory. Considering human movement as a dynamical process, estimating its dynamical system is non-trivial and would require us to completely observe all system variables (all joints), due to large degrees of freedom and complex interactions between the joints. In applications requiring inference of higher-level features from movement, the classical approach has been to seek certain invariant representations of the dynamical process that can still be determined from partial observations \cite{wolf1985determining}, for e.g. estimating the dynamics of human movement from a few skeletal joint sequences \cite{bissacco2005modeling,venkataraman2016shape}. However, such topological features are not predictive enough for complex datasets. More recently, with the availability of large datasets and neural networks, we are able to implicitly learn the dynamics much better, leading to significantly higher performance on challenging benchmarks \cite{vemulapalli2014human,zhang2019view,yan2018spatial,tang2018deep} but they tend to be very sensitive to changes in measurement settings during train and test times. 

\begin{figure*}[t]
\centering
\vspace{0.1in}
\includegraphics[width=0.9\textwidth]{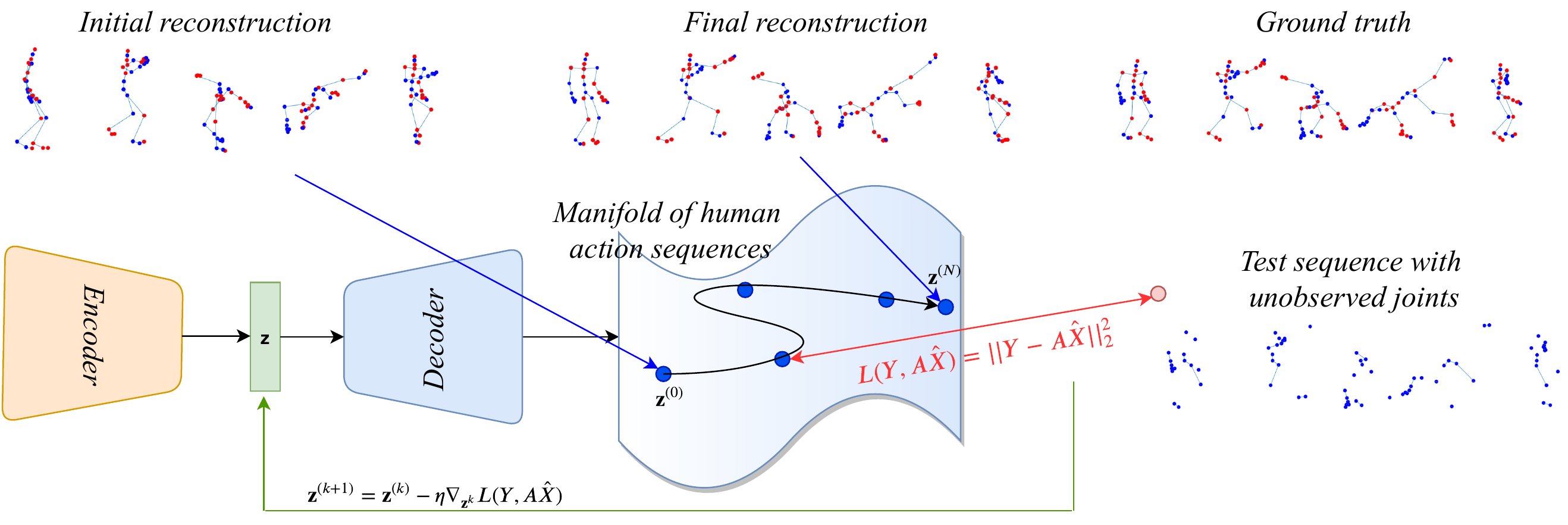}
\caption{\small{Block diagram illustrating the process of reconstructing dynamics of unobserved joints by projecting onto the manifold of human actions obtained as the range space of a generative model. For a test sequence with several missing joints, we get an initial estimate of the reconstruction using its encoder representation $\mathbf{z}^{(0)}$. Then, we optimize over the encoder representation space in order to minimize the distance between the output of the decoder and the given test sequence, as shown. This procedure can correct for the mismatch between training and test sets and greatly improves the final reconstructed sequence.}}
\label{fig:block_diagram}
\vspace{-0.15in}
\end{figure*}

\paragraph{\textbf{Contributions:}}
%
In this paper, we focus on the problem of reconstructing trajectories of unmarked/unobserved markers for 3D human action sequences at test-time. That is, given a human skeletal action sequence with a fraction of joints unmarked throughout the action, can we accurately reconstruct the dynamics of those unmarked joints and use the resulting completed action for action recognition without any modifications to the classification architectures? We study this problem when a significant subset of the joints is missing at both training as well as test times, with different rates of dropped joints. In practice, this can occur due to factors such as occlusions, faulty sensors, non-standard postures etc. In this setting, conventional deep learning frameworks fail due to the unaccounted distribution shift between training and testing data sets, and because when the joints are missing, a lot of information regarding the dynamics is lost. To this end,

\begin{enumerate}[nolistsep,leftmargin=*]

    \item We show that it is possible to effectively recover semantics and dynamics of human actions while having access to as few as $50\%$ of the joints during both training and testing.
    
    \item We pose the problem as an ill-posed linear inversion in the space of actions, and solve it with the help of a deep prior i.e., in the form of a pretrained temporal convolutional decoder. The solution to the inverse problem is obtained by approximately projecting the incomplete action onto the range of the decoder, as shown in Figure \ref{fig:block_diagram}.
    
    \item The proposed method is a test-time approach that can easily handle changes in the configuration and number of missing joints at test time, when compared to the training set. We demonstrate its effectiveness using real world mocap and Kinect datasets. We validate that the reconstructions recover information about the semantics and dynamics of the actions and show that in most cases the recovered actions perform within $~5\%$ of fully visible ground truth data in terms of action recognition on a pre-trained classifier, and yields a very low 3D joint reconstruction error of about 3cm.
\end{enumerate}

\section{Related Work}

\subsection{3D human action acquisition and completion}
The two most common methods of capturing 3D human skeletal actions are using motion capture (mocap) systems, and using depth-sensing cameras like Kinect and RealSense and then using a algorithm to estimate the joint locations. However, both these methods suffer from drawbacks. Not all joints may be recorded mainly due to occlusions and non-standard body types/poses. Moreover, in order to reduce the cost of mocap systems, we would like to employ fewer cameras which leads to the possibility of joints being unseen throughout the actions, which we refer to as unobserved joints. When the final application is action classification, it may be possible to classify directly with fewer joint trajectories, but this suffers from two disadvantages: (a) it is impractical to train a classifier for every combination of unobserved joints (b) the recognition performance is very poor when many joints are missing such as those considered in this paper. As we will show, we can get excellent performance when we first reconstruct the complete set of joint trajectories by exploiting information about the nature of human actions from a training dataset, and then performing the classification using a single pretrained classifier. Earlier works have considered the problem of human action completion in different ways -- human action prediction where given a few frames of a human action sequence, the future frames can be predicted employing machine learning algorithms \cite{barsoum2018hp,ghosh2017learning,Hernandez_2019_ICCV}, human motion synthesis \cite{fragkiadaki2015recurrent} and editing \cite{holden2016deep}, and pose estimation \cite{wang2019learning}. There are also traditional methods for human action completion using $k$-nearest neighbors from the training set \cite{aristidou2018self}, and matrix factorization methods \cite{gloersen2016predicting}. More recently, Yang et al. \cite{yang2019spatio} and \cite{xia2018nonlinear} propose using low-rank and sparsity priors to model spatial-temporal correlation for 3D human motion recovery problems. Here, the authors enforce the low-rankness of a matrix whose columns are the frames of a test sequence. However, as is in this paper, if the joint trajectories are missing throughout the action, the matrix is already low-rank and the optimization cannot yield the solution. 
In this paper, we consider the problem of predicting complete joint trajectories given a subset of joints of every frame in the sequence and propose a deep learning-based solution. Kucherenko et al. \cite{kucherenko2018neural} propose feed-forward pass through trained a LSTM autoencoder as a way of reconstructing human actions. We use a baseline similar to this method in our experiments.

\subsection{Deep priors for inverse problems} 
Several important problems in imaging can be cast as ill-posed linear inverse problems. The forward operation is given by $\mathbf{y} = A \mathbf{x}$, where $\mathbf{y} \in \mathbb{R}^m, \mathbf{x} \in \mathbb{R}^n, m < n$. The goal in inverse imaging is to reconstruct $\mathbf{x}$ from $\mathbf{y}$ which is generally ill-posed by exploiting structure of the desired $\mathbf{x}$ known apriori. In recent years, the prior knowledge comes in the form of a deep generative model (autoencoders, variational autoencoder, generative adversarial networks etc.). The process of reconstruction reduces to the problem of finding the closest point on the range space of the generative model $\hat{\mathbf{x}}$ such that $A\hat{\mathbf{x}} \approx \mathbf{y}$. This idea also has theoretical guarantees as shown by Bora et al. \cite{bora2017compressed} and Shah and Hegde \cite{shah2018solving}. Bora et al. \cite{bora2018ambientgan} showed that, in some cases, the generative model can also be \textit{trained} using the noisy images. Recent papers apply this idea for time-series imputation \cite{luo2018multivariate,yoon2018gain}. However, these techniques are shown to work for simpler time-series classification problems using measurements where some information is available from all dimensions. Also related are the works of Litany et al. \cite{litany2018deformable} which uses graph convolutional deep priors for 3D shape completion, and that of Holden et al. \cite{holden2018robust} who propose a denoising neural network for frame-wise denoising of motion capture data. In contrast to these works, we consider human action sequences in which certain joints are \textbf{completely unobserved} i.e., many dimensions of the time-series are missing. Thus, interpolation techniques in the time domain are not applicable. Also, current methods for skeleton completion is not usually performed at the level of actions (such as Kinect), and \textbf{do not take long-range dynamics into account}, which is the goal of this paper and necessary for human action recognition. We use a baseline very similar to \cite{holden2018robust} which shows the importance of modeling actions rather than individual frames. Also, unlike previous literature, we consider the effect of recovering trajectories of unobserved joints on the \textbf{action recognition} performance on a pre-trained classifier, and importantly, the setting where the training set for the autoencoder itself consists of actions with unobserved joints.

\subsection{Dynamical systems approach to deal with missing dimensions} 
In dynamical systems theory, the notion of reconstructing high-dimensional state-spaces from low-dimensional observations has been well-studied. For classical state-estimation approaches to apply, one often needs to make simplifying assumptions for state-dynamics, such as Markovian and linear dynamics. Such assumptions are not always reflective of the complexity of the task at hand -- reconstructing human action sequences. Another approach is to avoid such assumptions, but use methods from non-linear dynamics \cite{ljung2001system} to estimate surrogate state-spaces. From standard methods in non-linear dynamics, these surrogate state-spaces are only topologically equivalent to the true state-spaces, and do not have enough predictive information for high-level inference \cite{venkataraman2016shape}.

\section{Reconstructing dynamics of unmarked joints as an ill-posed linear inverse problem}
Recovering the dynamics of unobserved joints is an ill-posed inverse problem since we only have access to partial information of the activity. As we will show, this can be considered analogous to the inverse problems in imaging such as super-resolution, image inpainting or compressive sensing. However, unlike inverse imaging problems, it is not clear what kinds of priors work for human actions. We argue that a deep prior learned from a dataset of human actions acts as a good approximation to the space of all possible dynamical systems for human actions. As a result, we are able to implicitly constrain the dynamics of the recovered actions by restricting the solution to lie on the \emph{action manifold}. We formalize these ideas next.

Let the total number of joints per frame be denoted by $J$ and the number of frames per action sequence by $N$. Each joint is described by its 3D co-ordinates in space. Thus, by vectorizing, each skeleton can be represented by $3J-$dimensional vector and by stacking the $N$ frames in columns, we represent the human action as a matrix $X$ of size $3J \times N$. Let the number of joints observed be $K$, so the number of unobserved joints is $J-K$. The measurement operator $A$ then is a sub-sampling operator which drops $3(J-K)$ rows of $X$ to give us the observed action $Y$. As $J-K$ joints are unobserved, there are $3(J-K)$ rows of $Y$ which are unknown and we replace them with $0$ before further processing. Given $Y$, our eventual goal is to classify the action. As an intermediate step, we first reconstruct $\hat{X}$ from $Y$ which is the main focus of this paper. Clearly, this is an ill-posed linear inverse problem. The advantage of viewing it as such helps us adapt algorithms designed for inverse imaging problems such as image inpainting \cite{pathak2016context}, super-resolution \cite{dong2015image,ledig2017photo} and compressive imaging \cite{kulkarni2016reconnet}. In this paper, we adapt the most recent approaches based on generative priors \cite{bora2017compressed,bora2018ambientgan,ulyanov2018deep}. We note that the advantage of these methods over other methods in inverse imaging such as purely data-driven approaches \cite{dong2015image,kulkarni2016reconnet}, and unrolled iterative methods \cite{rick2017one,metzler2017learned,zhang2018ista} is that there is no requirement of paired $Y$ and $X$ for training. Thus, once we have a deep prior for human action sequences, the problem of reconstructing dynamics of unobserved joints can be solved using an optimization problem such that the output of the generative model $\hat{X}$ is closest (in some predefined sense) to the test sequence under consideration $Y$. 

\section{Learning the manifold of 3D human actions}
\label{sec:generative_model}
In order to approximate space of human actions, we employ a temporal convolutional autoencoder architecture to construct the generative model of human action sequences. We choose an autoencoder over generative adversarial networks \cite{goodfellow2014generative} or variational autoencoders \cite{kingma2013auto} because -- (1) autoencoders are much easier to train compared to the other frameworks and (2) the purpose of using the generative model in this paper is to perform reconstruction of test sequences rather than sampling new actions, which, as we will show, can be readily performed using an autoencoder. We can also employ more complex deep priors which can better model spatial relationships using graphs as recently shown by Wang et al. \cite{wang2019spatio}, however this is not the focus of this paper. Note that the term generative model is more broadly used here. Even though we cannot easily sample new actions using an autoencoder, once an initial valid latent space is given, we can move in the latent space in order to refine the output, based on some metric. We discuss this aspect in detail in Section \ref{sec:proj}.

\subsection{Autoencoder architecture}
\label{sec:ae}
As the generative model, we employ a temporal convolutional autoencoder. Both the encoder ($E$) and decoder ($D$) consists of a series of 1D convolutional layers operating in the temporal domain with ReLU non-linearity. After every convolution, we use average pooling to reduce the number of frames by half. We then use a fully-connected (FC) layer which produces the encoded/latent representation of the action, denoted by $\mathbf{z}$. The decoder reverses these operations with a series of transposed convolutional layers. The networks are trained using full/complete actions with access to information of all joint trajectories. The network is trained to minimize the Euclidean loss between the input sequence and the output of the decoder: $L(X, \hat{X}) = \sum_{n=1}^N \sum_{j=1}^J \norm{X_{n,j} - \hat{X}_{n,j}}_2^2$, where $X_{n,j}$ refers to the $j^{th}$ 3D joint location in the $n^{th}$ frame of the sequence. Other training details are provided in the appendix. We note that we can add an additional \textbf{adversarial loss} term to the above loss function in order to make the actions more realistic \cite{pathak2016context}. However, in our experiments, we did not observe any significant improvements using this additional loss term.

\subsection{Training the autoencoder with partially observed joint sequences}
\label{sec:ambientae}
The autoencoders in Sections \ref{sec:ae} are trained using complete actions with all joint sequences fully observed, $X$. However, complete actions may not be available at the training stage. \textit{An important contribution of this paper is showing that we can construct the manifold of human actions by training solely on action sequences with only a subset of the joints observed, $Y$.} We later show that this protocol leads to superior reconstruction compared to training with full actions. To this end, we modify the loss function as follows. The forward operator $A$ is the sampling operator which has the effect of dropping a subset of the joints. Using the knowledge of $A$, we use a masked loss function between the ground-truth measured sequence $Y = AX$ and the reconstructed sequence $\hat{X}$: $L_(Y, \hat{X}) = \sum_{n=1}^N \sum_{j=1}^J \norm{Y_{n,j} - A\hat{X}_{n,j}}_2^2$. The network architectures and the training protocols are identical to those trained using fully observed action sequences.

\section{Reconstruction via approximate projection onto the action manifold}
Once we have the deep prior model, in our case the decoder stage of the autoencoder, the training process is complete. The next step is to use the prior model in order to reconstruct the trajectories of unobserved joints given an incomplete action sequence. To this end, we propose to project the incomplete action to the range space of the generator, which ideally is the same as the manifold of complete human action sequences. 

\paragraph{Initialization: Feed-forward pass through the trained autoencoder:}
As a baseline method, we can simply feed the incomplete action sequence through the autoencoder and use the output of the decoder as the reconstruction. This is expected to fail, because of the distribution shift between training and test set, especially in the case of the autoencoder trained with complete actions. However, in the case of the autoencoder trained on subsets of joint trajectories, even this simple method can provide a reasonable reconstructed sequence. This is used as initialization for the optimization algorithm below, $\mathbf{z}^{(0)}$.

\subsection{Optimizing the latent representation}
\label{sec:proj}
We can further improve the reconstruction quality from above by directly optimizing the encoded/latent representation, $\mathbf{z}$, such that the Euclidean distance between the reconstructed action sequence and the input incomplete sequence. This method is inspired by Bora et al. \cite{bora2017compressed} where the authors propose this method for inverse problems in imaging. The optimization problem is given by 

\begin{equation}
    \mathbf{z}^* = \arg\min_\mathbf{z} \norm{Y - AD(\mathbf{z})}_2^2, \quad \hat{X} = D(\mathbf{z}^*). \label{eqn:proj}
\end{equation}

This ameliorates the train/test distribution-shift issue and allows a better exploration of the output space of the decoder. We solve this optimization problem using gradient descent-type method. As the optimization problem is non-convex, the optimization returns a $\mathbf{z}^*$ which is guaranteed to be only locally optimal. Empirically, we find that the solutions obtained using this procedure provide high quality reconstructions. 

\section{Measuring reconstruction performance}
Our main goal in this paper is to recover trajectories of unobserved joints from incomplete human actions. We choose the following three methods for measuring the quality of reconstructed actions.

\subsection{Reconstruction error}
We measure the root-mean-square error (RMSE) between the reconstructed actions and the ground-truth actions. For a reconstructed sequence $\hat{X}$ and the corresponding ground-truth sequence $X$, RMSE is given by $RMSE(X, \hat{X}) = \sqrt{\frac{1}{NJ}\sum_{n=1}^N \sum_{j=1}^J \norm{X_{n,j} - \hat{X}_{n,j}}_2^2}$.

\subsection{Classification performance} 
\label{sec:classifier}
An important reason for reconstructing action sequences is to employ predefined classification pipelines without any modification. Therefore, we train a single action classifier on sequences with information of all joints, feed the reconstructed sequences as test inputs, and use the classification performance as a metric for quality of reconstructed actions.

\vspace{-0.1in}
\paragraph{Classifier architecture:} In all our experiments, we employ a simple popular architecture for 3D human action recognition based on temporal convolutional networks (TCNs) \cite{kim2017interpretable}. The classifer consists of a series of temporal convolutional blocks. Each block consists of layers of 1D convolutional layers operating in the temporal domain with ReLU non-linearity. We employ batch normalization for each layer. Residual connections are employed from one block to the next. After every block, average pooling is employed to reduce the number of frames by half. Finally a fully-connected (FC) layer with softmax is used to map to a probability distribution over the classes. As there may be a domain shift between the original actions and the reconstructed actions from the autoencoder, we train a single classifier on complete action sequences. Other training details are provided in the appendix. We also visualize the effectiveness of the reconstructions for downstream applications, like classification, with t-SNE embeddings in 2D \cite{maaten2008visualizing}. We use the feature maps of the penultimate layer of the trained classifier and are shown in the appendix.


\section{Experimental results}
In the experiments section, we simulate mocap and Kinect with unmarked/unobserved joints using benchmark datasets which have fully observed joints.

\subsection{Baselines}
\label{sec:baselines}
\vspace{-0.1in}
\paragraph{Dictionary learning and sparse coding \cite{mairal2009online,scikit-learn}:} This is a commonly used algorithm for solving ill-posed problems. We first construct a dictionary based on all the action sequences in the training set. Dictionary learning is performing by using matrix factorization with sparsity and norm constraints on the matrix factors. Given a test sequence, in which joint trajectories are missing, we express the test sequence as a sparse linear combination of the atoms in the dictionary. The linear combination is used as the completed sequence. The sparse code is found using orthogonal matching pursuit. Please see the appendix for details. 

\vspace{-0.15in}
\paragraph{Frame-wise completion with deep learning:} In order to demonstrate the effect of modeling temporal variations (which is one of the main contributions of this paper), jointly with purely frame-wise information, we construct a baseline very similar to that in Section \ref{sec:generative_model}. The only difference is that instead of encoding and decoding actions directly, the action is split into frames and the decoding is done on a per-frame basis. Thus, no temporal correlations are modeled in the deep prior. Further details about the architecture are given in the appendix. This baseline is similar in spirit to the work of Litany et al. \cite{litany2018deformable} for 3D human shape completion.

\vspace{-0.1in}
\paragraph{Decoder reconstruction:} As described in Section \ref{sec:ambientae}, a feed-forward pass through the trained autoencoder itself can be used to complete the sequence. And we use this as a baseline to compare with the further latent space optimization we propose in this paper.

\subsection{HDM05 mocap dataset \cite{cg-2007-2}}
\paragraph{Dataset details:} HDM05 is a large publicly available and challenging database of 3D human actions with $2337$ action samples. There are $130$ different types of actions performed by $5$ subjects and recorded in a laboratory setting using an optical motion capture system. Each skeleton is made up of $31$ joints. For our experiments, we resample all the actions such that the length of the every action sequence $N = 100$. Thus $X, Y, \hat{X} \in \mathbb{R}^{93 \times 100}$. All sequences are normalized so that the hip joint is fixed at the origin in 3D space. We perform $5$-fold cross-validation. For each run, we use 4 subjects for training and the remaining subject for testing with about $1850$ samples for training and the rest for testing. Note that, in all cases, as a pre-processing step, the unobserved joint trajectories are replaced with the respective mean trajectories computed over the observed joints in the training set. However, simply using zeros as the initialization for the missing joints produces nearly identical results and these are shown in the appendix.

\vspace{-0.1in}
\paragraph{Network architectures:}
The encoder consists of 4 temporal convolution layers with filter size of 4, and the number of feature maps in each layer is set to $75$ (equal to the number of channels at the input layer). We use a latent space dimension of $200$. The decoder consists of 4 temporal transposed convolutional layers. For our experiments, we train multiple autoencoders with actions consisting of random subsets of joint sequences sampled from the actions. Different fractions of joints included for training each autoencoder: $100\%, 75\%, 50\%$. We will use the term Observed-to-Total Percentage (OTP) = $\frac{K}{J} \times 100$, to denote this quantity. For classification, we use a TCN classifier similar \cite{kim2017interpretable}. It consists of 3 TCN blocks with one convolutional layer each.

\begin{figure*}[!ht]
\centering
\includegraphics[trim={4cm 8cm 4cm 0},clip,width=0.8\linewidth]{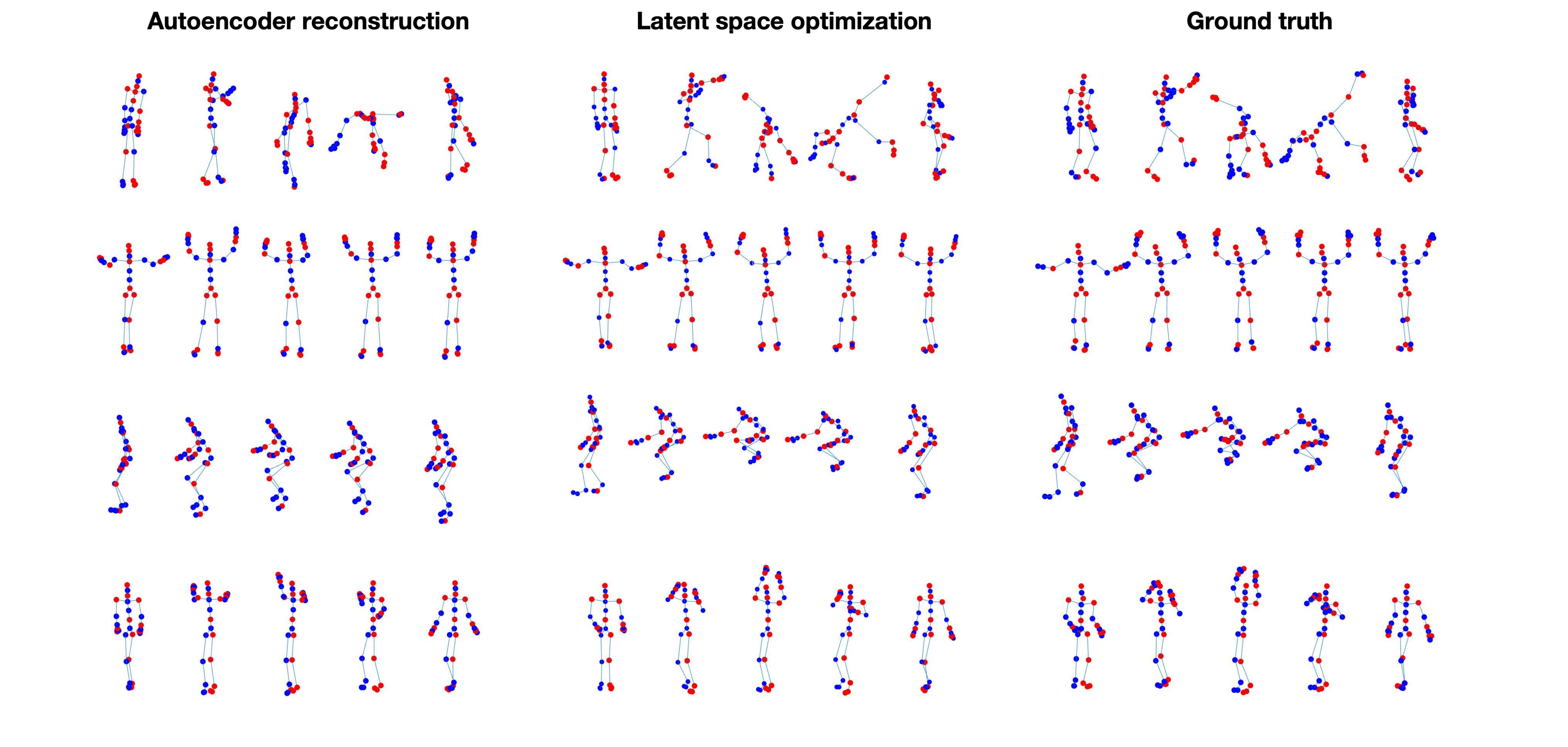}
\vspace{-0.25in}
\caption{\small{Reconstructed actions for the HDM05 database for train OTP / test OTP = $75/50$. From the top row, the actions shown are ``Cartwheel" ,``Hand waving" and ``Grab low". The first column shows the reconstructions obtained by a simple feedforward pass through the trained autoencoder: $D(E(Y))$. The middle column shows the output of the proposed approach which solves the optimization problem in Equation \ref{eqn:proj}: $D(\mathbf{z}^*)$. Blue dots represent the observed joints and red dots represent the unobserved. We clearly observe that the optimization approach produces superior reconstructions.}}
\label{fig:hdm_results}
\end{figure*}

\begin{table*}[htb]
\small
\centering
\begin{tabular}{c|c|c|c|c|c}
Train OTP / Test OTP & Method & \multicolumn{2}{c|}{HDM} & \multicolumn{2}{c}{NTU} \\
\cline{3-6}
& & RMSE (cm) & Acc (\%) & RMSE (cm) & Acc (\%) \\
\hline

100/100 & $D(E(Y))$ & 3.48 & 79.23 & 2.67 & 74.77\\
\hline


\multirow{2}{*}{100/75} & $D(E(Y))$ & 6.21 & 61.21 & 6.18 & 55.90\\
& $D(\mathbf{z}^*)$ & \textbf{2.18} & \textbf{78.04} & \textbf{3.33} & \textbf{72.26}\\
\hline

\multirow{2}{*}{100/50} & $D(E(Y))$ & 9.87 & 28.99 & 9.39 & 31.89 \\
& $D(\mathbf{z}^*)$ & \textbf{2.99} & \textbf{73.47} & \textbf{5.19} & \textbf{65.15} \\
\hline




\multirow{2}{*}{75/75} & $D(E(Y))$ & 6.07 & 68.72 & 6.08 & 59.15 \\
& $D(\mathbf{z}^*)$ & \textbf{2.27} & \textbf{78.61} & \textbf{3.49} & \textbf{71.64} \\
\hline

\multirow{2}{*}{75/50} & $D(E(Y))$ & 8.52 & 44.81 & 9.11 & 37.38 \\
& $D(\mathbf{z}^*)$ & \textbf{2.98} & \textbf{74.71} & \textbf{5.30} & \textbf{64.81} \\
\hline

\multirow{2}{*}{50/50} & $D(E(Y))$ & 8.77 & 43.55 &9.37 & 34.93 \\
& $D(\mathbf{z}^*)$ & \textbf{3.05} & \textbf{74.37} & \textbf{5.29} & \textbf{64.59} \\
\end{tabular}
\caption{\small{Experimental results for HDM05 (averaged over 5 folds) and NTU datasets for varying train/test OTPs in terms of RMSE and action recognition accuracy (Acc). We observe easily that the proposed optimization-based reconstruction is far superior to a feedforward pass through the autoencoder in terms of all the metrics considered. As the train OTP is reduced, performance degrades more gracefully in the case of the optimization-based approach. In all cases, we can get to within $5 \%$ points of the oracle action recognition performance (train /test OTP = $100/100$).}}
\vspace{-0.15in}
\label{table:hdm_ntu_results_1}
\end{table*}

\begin{table}[t]
\small
\renewcommand{\tabcolsep}{0.5mm}
\centering
\begin{tabular}{c|c|c|c|c}
Method & \multicolumn{2}{c|}{HDM} & \multicolumn{2}{c}{NTU} \\
\cline{2-5}
& RMSE (cm) & Acc (\%) & RMSE (cm) & Acc (\%) \\
\hline
Sparse Coding \cite{mairal2009online} & 20.27 & 11.30 & 19.32 & 7.22\\
\hline
\makecell{Frame-wise \\ \makecell{$D_F(E_F(Y_F))$ \\ similar to \cite{holden2018robust}}} & 21.47 & 12.30 & 20.40 & 9.69 \\
\hline
\makecell{Frame-wise \\ $D_F(\mathbf{z_F}^*)$} & 21.28 & 13.44 & 20.19 & 8.62 \\
\hline
Action $D(E(Y))$ & 8.52 & 44.81 & 9.11 & 37.38 \\
\hline
\makecell{Action $D(\mathbf{z}^*)$ \\ (Proposed)} & \textbf{2.98} & \textbf{74.71} & \textbf{5.30} & \textbf{64.81} \\
\end{tabular}
\vspace{-0.1in}
\caption{\small{Experimental results for HDM05 (averaged over 5 folds) and NTU datasets compared to different baselines for train/test OTP = 75/50. We observe easily that the proposed optimization-based reconstruction is superior to all the baselines considered. $D_F$ and $E_F$ refer to the fact that the encoder and decoder operate on a single frame at a time, rather than an action sequence.}}
\label{table:hdm_ntu_results_2}
\end{table}

\vspace{-0.1in}
\paragraph{Reconstruction performance and visualization:}
We compare the results of the proposed method with the baselines described in Section \ref{sec:baselines}. As explained in Section \ref{sec:proj}, using our proposed method, we can achieve significantly better reconstructions by using an optimization procedure over the latent space of the autoencoder model in order to minimize the Euclidean distance between the reconstruction and input test sequence with only a subset of observed joint trajectories. Note that the parameters of the generative model, the decoder $D$ in our case, are held fixed for this optimization. We use Adam optimizer for $500$ iterations with an initial learning rate of $1.0$. As the initialization, we use the latent representation of the incomplete action obtained by a feed-forward pass through the autoencoder. We carry out reconstruction experiments on the test set for autoencoders trained with different fractions of observed joints. 

In order to better test the generalization ability, we use test-time OTPs that are different from training-time OTPs. For the reconstructed sequences thus obtained, we use a pre-trained classifier to classify the test set reconstruction into one of the pre-defined 130 classes for different variants and test-time OTPs of the autoencoders. The RMSE of the reconstructed sequences as well as the action recognition performance, averaged over the five folds, are shown in Tables \ref{table:hdm_ntu_results_1} and \ref{table:hdm_ntu_results_2}, and a few sample reconstructed skeletal sequences are shown in Figure \ref{fig:hdm_results}. 

\noindent \textbf{Results:} We observe in all cases where either the training or test OTP is less than $100 \%$, the proposed method of solving the optimization problem in Equation \eqref{eqn:proj} i.e., using $D(\mathbf{z}^*)$ leads to significantly better results compared to baselines including using a single feed-forward pass through the autoencoder i.e., $D(E(Y))$, where $D$ and $E$ are the decoder and encoder respectively. In almost all cases, using the optimization approach yields accuracies within $5 \%$ points of the oracle action recognition performance and about 3 cm error in terms of RMSE. We also observe for the cases with training OTP = $100 \%$, using a different test OTP causes more degradation in performance than when the train OTP = $75 \%, 50 \%$. \textbf{We note that we also tested two additional simpler baselines}: (1) perform reconstruction of skeletons per frame by replacing every unobserved joint with the closest observed joint in the skeleton, and then use the pre-trained classifier and (2) train the classifier directly on the action sequences with only a subset of joints observed. Irrespective of the train/test OTP, both these baselines fail and yield only accuracies which are close to chance. 

In the appendix, we provide an additional baseline using a denoising autoencoder (DAE) framework where the only difference from the framework presented here is that, we assume that we have access to a training set with all joints trajectories available and the DAE maps from incomplete actions to complete ones. That is, rather than measuring the error only in the subset of joints (Equation 2), we measure the error for all the joints as the ground-truth training actions are assumed to be perfect. The results show this extra supervision actually hurts the final test performance when the Train and Test OTPs are different, compared to the proposed method. As in the other cases, latent space optimization helps improve the performance. We also performed an experiment where, different joints are missing for different frames, rather than joints missing for the entire sequence. The proposed method can be applied without any modifications, and the results show that the performance and trends remains the same as in the case of completely missing joints (which is the focus of this paper). These results are also provided in the appendix.

\vspace{-0.1in}
\paragraph{Reconstruction using structured masks:} In the above, we trained and tested autoencoders with random subsets of joints dropped. In this experiment, we drop joints in a structured fashion during test time. We carry out four sets of experiments with the joints corresponding to the following body parts dropped: right arm (6 joints), left arm (6), right leg (4) and left leg (4). This demonstrates how occlusion of different limbs can affect the performance of our algorithm. Note that the autoencoders were trained on random subsets of joints, as before. The results are shown in Table \ref{table:hdm_results_limb}. We see once again that the proposed algorithm yields good recognition performance compared to the baselines considered. These results also illustrate another important aspect of the work---even though the train and test OTP are nearly the same as in the case of occluding right/left leg, the baseline method's performance suffers significantly as the distribution of the joints being dropped is very different from the training scenario (random subsets of joints are dropped), and the proposed latent space optimization technique allows better exploration of the decoder space in order to find better reconstructions.

\vspace{-0.1in}
\paragraph{Generalizing to unseen actions:} 
Here, instead of having a test set having actions from an unseen subject, the test set actions also belong to \textit{classes} which are unseen during training time. For our experiment, we use a random split of $100$ and $30$ action classes for the train and test set respectively. The joint RMSE results shown in Table \ref{table:actions_split} show that our method is indeed able to yield good results in this challenging setting, showing that the proposed method is able to generalize to both unseen subjects and unseen classes.

\paragraph{}

\begin{table}[t]
\small
\renewcommand{\tabcolsep}{1mm}
\centering
\begin{tabular}{c|c|c|c|c}
\makecell{Occluded limb/ \\ Frac. joints dropped} & \makecell{Train \\ OTP} & Method & \makecell{RMSE \\ (cm)} & \makecell{Recog. \\ Acc. (\%)} \\
\hline
\multirow{4}{*}{\makecell{Right arm/ \\ $6/31 \approx 20\%$}} & \multirow{2}{*}{90} & $D(E(Y))$ & 9.66 & 48.97\\
& & $D(\mathbf{z}^*)$ & \textbf{7.03} & \textbf{61.55} \\
\cline{2-5}
& \multirow{2}{*}{75} & $D(E(Y))$ & 9.43 & 51.62\\
& & $D(\mathbf{z}^*)$ & \textbf{6.54} & \textbf{64.89} \\
\hline
\multirow{4}{*}{\makecell{Left arm/ \\ $6/31 \approx 20\%$}} & \multirow{2}{*}{90} & $D(E(Y))$ & 9.51 & 51.53\\
& & $D(\mathbf{z}^*)$ & \textbf{6.88} & \textbf{65.29} \\
\cline{2-5}
& \multirow{2}{*}{75} & $D(E(Y))$ & 9.38 & 52.29\\
& & $D(\mathbf{z}^*)$ & \textbf{6.55} & \textbf{68.49}\\
\hline
\multirow{2}{*}{\makecell{Right leg/ \\ $4/31 \approx 13\%$}} & \multirow{2}{*}{90} & $D(E(Y))$ & 6.32 & 58.61\\
& & $D(\mathbf{z}^*)$ & \textbf{4.25} & \textbf{67.95}\\
\hline
\multirow{2}{*}{\makecell{Left leg/ \\ $4/31 \approx 13\%$}} & \multirow{2}{*}{90} & $D(E(Y))$ & 6.50 & 53.30\\
& & $D(\mathbf{z}^*)$ & \textbf{4.24} & \textbf{64.20}\\
\end{tabular}
\caption{Average classification results (over 5 folds) of reconstructed actions on the HDM05 database. The inputs are actions with contiguous body parts that are hidden or unobserved. Here, $D(E(Y))$ is the baseline and $D(\mathbf{z}^*)$ is the proposed optimization strategy.}
\label{table:hdm_results_limb}
\vspace{-0.25in}
\end{table}

\begin{table}[htb]
\small
\centering
\begin{tabular}{c|c|c}
Train OTP / Test OTP & Method & RMSE (cm) \\
\hline

100/100 & $D(E(Y))$ &  1.79 \\
\hline

\multirow{2}{*}{100/75} & $D(E(Y))$ & 5.34\\
& $D(\mathbf{z}^*)$ & \textbf{1.43} \\
\hline

\multirow{2}{*}{100/50} & $D(E(Y))$ & 9.42 \\
& $D(\mathbf{z}^*)$ & \textbf{2.14} \\
\hline

\multirow{2}{*}{75/75} & $D(E(Y))$ & 5.55 \\
& $D(\mathbf{z}^*)$ & \textbf{1.99} \\
\hline

\multirow{2}{*}{75/50} & $D(E(Y))$ & 6.92\\
& $D(\mathbf{z}^*)$ & \textbf{2.68} \\
\hline

\multirow{2}{*}{50/50} & $D(E(Y))$ & 6.67\\
& $D(\mathbf{z}^*)$ & \textbf{2.76} \\
\end{tabular}
\caption{\small{Experimental results for HDM05 when training and test action classes are different, for varying train/test OTPs in terms of RMSE. The results are averaged over 5 runs where for each run, we randomly select the action classes for train and test. We observe easily that the proposed optimization-based reconstruction is far superior to a feedforward pass through the autoencoder.}}
\vspace{-0.15in}
\label{table:actions_split}
\end{table}

\vspace{-0.2in}
\subsection{NTU RGB-D dataset \cite{shahroudy2016ntu}}
\vspace{-0.1in}
\paragraph{Dataset details:} This is a large database of about $56000$ 3D skeletal action sequences obtained from Kinect of actions belonging to $60$ classes and performed by $45$ subjects. For each skeleton, $25$ joint locations are provided. We resample all the sequences to have $N = 50$ frames. Thus $X, Y, \hat{X} \in \mathbb{R}^{75 \times 50}$. We perform the experiments in the cross-subject setting and use the train-test split as suggested by the authors of the dataset. The training set consists of about $40000$ examples and the remaining are in the test set. All sequences are normalized so that the hip is fixed at the origin in 3D space.

\vspace{-0.1in}
\paragraph{Network architectures:}
The generative mode is a temporal convolutional autoencoder. The encoder consists of 3 temporal convolution layers with filter size of $8$, and the number of feature maps in each layer is set to $75$ (equal to the number of channels at the input layer). We use a latent space dimension of $200$. The decoder consists of 3 temporal transposed convolutional layers. As the action classifier, we use a TCN classifier identical to that proposed by Kim and Reiter \cite{kim2017interpretable}. It consists of 3 TCN blocks with 3 convolutional layers each.

\vspace{-0.1in}
\paragraph{Reconstruction performance:} We conduct an identical set of experiments as in the case of the HDM05 dataset. The action classfication accuracies and RMSE are shown in Tables \ref{table:hdm_ntu_results_1} and \ref{table:hdm_ntu_results_2}. Skeletal visualizations are provided in the appendix. The trends observed are the same as those in HDM05. Compared to all the baselines considered including (1) sparse coding (2) frame-wise completion and (3) using the autoencoder reconstruction, the proposed method of latent space optimization achieves far superior results especially when train and test OTPs are considerably different, and gets close to oracle classification performance and RMSE even with just $75 \%$ of observed joints. Results for other train/test OTPs are provided in the appendix.

\vspace{-0.1in}
\section{Conclusions}
We consider the problem of reconstructing completely unobserved dimensions of a multi-variate time series. The problem is traditionally studied in the framework of system identification and non-linear dynamics. However, for tractability, such methods make strong assumptions on the data such as linearity of the underlying dynamical system, sparsity of observations in transform domains etc. In this paper, we study the specific example of reconstructing dynamics of unobserved joints from 3D human actions, for which we cannot easily construct hand-crafted priors. We propose to first construct a deep prior of complete actions, even when the training data has up to $50\%$ of the joints missing. The reconstruction problem then can be solved by projecting the observed action with missing joints onto the action manifold, which is done via optimization in the latent space. Through extensive experiments and different metrics, we show that the proposed approach can effectively recover the dynamics of unobserved joints. Interesting extension of this ideas for human actions include designing stronger priors using spatio-temporal graph convolutional autoencoders and different improved loss functions for latent space optimization.

{\small
\bibliographystyle{ieee_fullname}
\bibliography{egpaper}

\begin{thebibliography}{10}\itemsep=-1pt

\bibitem{KinectTherapy}
Kinect'in therapy: Disabled access to the {Xbox Kinect} - hardware and games.
\newblock \url{http://kinect-therapy.blogspot.uk}.
\newblock Accessed: 2020-03-03.

\bibitem{aristidou2018self}
Andreas Aristidou, Daniel Cohen-Or, Jessica~K Hodgins, and Ariel Shamir.
\newblock Self-similarity analysis for motion capture cleaning.
\newblock In {\em Computer Graphics Forum}. Wiley Online Library, 2018.

\bibitem{barsoum2018hp}
Emad Barsoum, John Kender, and Zicheng Liu.
\newblock {HP-GAN}: Probabilistic {3D} human motion prediction via {GAN}.
\newblock In {\em Proceedings of the IEEE Conference on Computer Vision and
  Pattern Recognition Workshops}, pages 1418--1427, 2018.

\bibitem{bissacco2005modeling}
Alessandro Bissacco.
\newblock Modeling and learning contact dynamics in human motion.
\newblock In {\em 2005 IEEE Computer Society Conference on Computer Vision and
  Pattern Recognition (CVPR'05)}, volume~1, pages 421--428. IEEE, 2005.

\bibitem{bora2017compressed}
Ashish Bora, Ajil Jalal, Eric Price, and Alexandros~G Dimakis.
\newblock Compressed sensing using generative models.
\newblock In {\em Proceedings of the 34th International Conference on Machine
  Learning-Volume 70}, pages 537--546. JMLR. org, 2017.

\bibitem{bora2018ambientgan}
Ashish Bora, Eric Price, and Alexandros~G Dimakis.
\newblock Ambient{GAN}: Generative models from lossy measurements.
\newblock {\em ICLR}, 2:5, 2018.

\bibitem{dong2015image}
Chao Dong, Chen~Change Loy, Kaiming He, and Xiaoou Tang.
\newblock Image super-resolution using deep convolutional networks.
\newblock {\em IEEE transactions on pattern analysis and machine intelligence},
  38(2):295--307, 2015.

\bibitem{fragkiadaki2015recurrent}
Katerina Fragkiadaki, Sergey Levine, Panna Felsen, and Jitendra Malik.
\newblock Recurrent network models for human dynamics.
\newblock In {\em Proceedings of the IEEE International Conference on Computer
  Vision}, pages 4346--4354, 2015.

\bibitem{ghosh2017learning}
Partha Ghosh, Jie Song, Emre Aksan, and Otmar Hilliges.
\newblock Learning human motion models for long-term predictions.
\newblock In {\em 2017 International Conference on 3D Vision (3DV)}, pages
  458--466. IEEE, 2017.

\bibitem{gloersen2016predicting}
{\O}yvind Gl{\o}ersen and Peter Federolf.
\newblock Predicting missing marker trajectories in human motion data using
  marker intercorrelations.
\newblock {\em PloS one}, 11(3):e0152616, 2016.

\bibitem{goodfellow2014generative}
Ian Goodfellow, Jean Pouget-Abadie, Mehdi Mirza, Bing Xu, David Warde-Farley,
  Sherjil Ozair, Aaron Courville, and Yoshua Bengio.
\newblock Generative adversarial nets.
\newblock In {\em Advances in neural information processing systems}, pages
  2672--2680, 2014.

\bibitem{Hernandez_2019_ICCV}
Alejandro Hernandez, Jurgen Gall, and Francesc Moreno-Noguer.
\newblock Human motion prediction via spatio-temporal inpainting.
\newblock In {\em The IEEE International Conference on Computer Vision (ICCV)},
  October 2019.

\bibitem{holden2018robust}
Daniel Holden.
\newblock Robust solving of optical motion capture data by denoising.
\newblock {\em ACM Transactions on Graphics (TOG)}, 37(4):1--12, 2018.

\bibitem{holden2016deep}
Daniel Holden, Jun Saito, and Taku Komura.
\newblock A deep learning framework for character motion synthesis and editing.
\newblock {\em ACM Transactions on Graphics (TOG)}, 35(4):138, 2016.

\bibitem{junejo2008cross}
Imran~N Junejo, Emilie Dexter, Ivan Laptev, and Patrick P{\'U}rez.
\newblock Cross-view action recognition from temporal self-similarities.
\newblock In {\em European Conference on Computer Vision}, pages 293--306.
  Springer, 2008.

\bibitem{kim2017interpretable}
Tae~Soo Kim and Austin Reiter.
\newblock Interpretable {3D} human action analysis with temporal convolutional
  networks.
\newblock In {\em 2017 IEEE conference on computer vision and pattern
  recognition workshops (CVPRW)}, pages 1623--1631. IEEE, 2017.

\bibitem{kingma2014adam}
Diederik~P Kingma and Jimmy Ba.
\newblock Adam: A method for stochastic optimization.
\newblock {\em arXiv preprint arXiv:1412.6980}, 2014.

\bibitem{kingma2013auto}
Diederik~P Kingma and Max Welling.
\newblock Auto-encoding variational bayes.
\newblock {\em arXiv preprint arXiv:1312.6114}, 2013.

\bibitem{kucherenko2018neural}
Taras Kucherenko, Jonas Beskow, and Hedvig Kjellstr{\"o}m.
\newblock A neural network approach to missing marker reconstruction in human
  motion capture.
\newblock {\em arXiv preprint arXiv:1803.02665}, 2018.

\bibitem{kulkarni2016reconnet}
Kuldeep Kulkarni, Suhas Lohit, Pavan Turaga, Ronan Kerviche, and Amit Ashok.
\newblock Reconnet: Non-iterative reconstruction of images from compressively
  sensed measurements.
\newblock In {\em Proceedings of the IEEE Conference on Computer Vision and
  Pattern Recognition}, pages 449--458, 2016.

\bibitem{ledig2017photo}
Christian Ledig, Lucas Theis, Ferenc Husz{\'a}r, Jose Caballero, Andrew
  Cunningham, Alejandro Acosta, Andrew Aitken, Alykhan Tejani, Johannes Totz,
  Zehan Wang, et~al.
\newblock Photo-realistic single image super-resolution using a generative
  adversarial network.
\newblock In {\em Proceedings of the IEEE conference on computer vision and
  pattern recognition}, pages 4681--4690, 2017.

\bibitem{litany2018deformable}
Or Litany, Alex Bronstein, Michael Bronstein, and Ameesh Makadia.
\newblock Deformable shape completion with graph convolutional autoencoders.
\newblock In {\em Proceedings of the IEEE conference on computer vision and
  pattern recognition}, pages 1886--1895, 2018.

\bibitem{ljung2001system}
Lennart Ljung.
\newblock System identification.
\newblock {\em Wiley Encyclopedia of Electrical and Electronics Engineering},
  2001.

\bibitem{luo2018multivariate}
Yonghong Luo, Xiangrui Cai, Ying Zhang, Jun Xu, et~al.
\newblock Multivariate time series imputation with generative adversarial
  networks.
\newblock In {\em Advances in Neural Information Processing Systems}, pages
  1596--1607, 2018.

\bibitem{maaten2008visualizing}
Laurens van~der Maaten and Geoffrey Hinton.
\newblock Visualizing data using t-{SNE}.
\newblock {\em Journal of machine learning research}, 9(Nov):2579--2605, 2008.

\bibitem{mairal2009online}
Julien Mairal, Francis Bach, Jean Ponce, and Guillermo Sapiro.
\newblock Online dictionary learning for sparse coding.
\newblock In {\em Proceedings of the 26th annual international conference on
  machine learning}, pages 689--696, 2009.

\bibitem{metzler2017learned}
Chris Metzler, Ali Mousavi, and Richard Baraniuk.
\newblock Learned {D-AMP}: Principled neural network based compressive image
  recovery.
\newblock In {\em Advances in Neural Information Processing Systems}, pages
  1772--1783, 2017.

\bibitem{cg-2007-2}
M. M\"{u}ller, T. R\"{o}der, M. Clausen, B. Eberhardt, B. Kr\"{u}ger, and A.
  Weber.
\newblock Documentation mocap database {HDM05}.
\newblock Technical Report CG-2007-2, Universit\"{a}t Bonn, June 2007.

\bibitem{pathak2016context}
Deepak Pathak, Philipp Krahenbuhl, Jeff Donahue, Trevor Darrell, and Alexei~A
  Efros.
\newblock Context encoders: Feature learning by inpainting.
\newblock In {\em Proceedings of the IEEE conference on computer vision and
  pattern recognition}, pages 2536--2544, 2016.

\bibitem{scikit-learn}
F. Pedregosa, G. Varoquaux, A. Gramfort, V. Michel, B. Thirion, O. Grisel, M.
  Blondel, P. Prettenhofer, R. Weiss, V. Dubourg, J. Vanderplas, A. Passos, D.
  Cournapeau, M. Brucher, M. Perrot, and E. Duchesnay.
\newblock Scikit-learn: Machine learning in {P}ython.
\newblock {\em Journal of Machine Learning Research}, 12:2825--2830, 2011.

\bibitem{regazzoni2014rgb}
Daniele Regazzoni, Giordano de Vecchi, and Caterina Rizzi.
\newblock Rgb cams vs rgb-d sensors: Low cost motion capture technologies
  performances and limitations.
\newblock {\em Journal of Manufacturing Systems}, 33(4):719--728, 2014.

\bibitem{rick2017one}
JH Rick~Chang, Chun-Liang Li, Barnabas Poczos, BVK Vijaya~Kumar, and Aswin~C
  Sankaranarayanan.
\newblock One network to solve them all--solving linear inverse problems using
  deep projection models.
\newblock In {\em Proceedings of the IEEE International Conference on Computer
  Vision}, pages 5888--5897, 2017.

\bibitem{shah2018solving}
Viraj Shah and Chinmay Hegde.
\newblock Solving linear inverse problems using {GAN} priors: An algorithm with
  provable guarantees.
\newblock In {\em 2018 IEEE International Conference on Acoustics, Speech and
  Signal Processing (ICASSP)}, pages 4609--4613. IEEE, 2018.

\bibitem{shahroudy2016ntu}
Amir Shahroudy, Jun Liu, Tian-Tsong Ng, and Gang Wang.
\newblock {NTU RGB+D}: A large scale dataset for 3{D} human activity analysis.
\newblock In {\em Proceedings of the IEEE conference on computer vision and
  pattern recognition}, pages 1010--1019, 2016.

\bibitem{tang2018deep}
Yansong Tang, Yi Tian, Jiwen Lu, Peiyang Li, and Jie Zhou.
\newblock Deep progressive reinforcement learning for skeleton-based action
  recognition.
\newblock In {\em Proceedings of the IEEE Conference on Computer Vision and
  Pattern Recognition}, pages 5323--5332, 2018.

\bibitem{ulyanov2018deep}
Dmitry Ulyanov, Andrea Vedaldi, and Victor Lempitsky.
\newblock Deep image prior.
\newblock In {\em Proceedings of the IEEE Conference on Computer Vision and
  Pattern Recognition}, pages 9446--9454, 2018.

\bibitem{vemulapalli2014human}
Raviteja Vemulapalli, Felipe Arrate, and Rama Chellappa.
\newblock Human action recognition by representing {3D} skeletons as points in
  a lie group.
\newblock In {\em Proceedings of the IEEE conference on computer vision and
  pattern recognition}, pages 588--595, 2014.

\bibitem{venkataraman2016shape}
Vinay Venkataraman and Pavan Turaga.
\newblock Shape distributions of nonlinear dynamical systems for video-based
  inference.
\newblock {\em IEEE transactions on pattern analysis and machine intelligence},
  38(12):2531--2543, 2016.

\bibitem{wang2019learning}
Chunyu Wang, Haibo Qiu, Alan~L Yuille, and Wenjun Zeng.
\newblock Learning basis representation to refine 3d human pose estimations.
\newblock In {\em Proceedings of the AAAI Conference on Artificial
  Intelligence}, volume~33, pages 8925--8932, 2019.

\bibitem{wang2019spatio}
He Wang, Edmond~SL Ho, Hubert~PH Shum, and Zhanxing Zhu.
\newblock Spatio-temporal manifold learning for human motions via long-horizon
  modeling.
\newblock {\em IEEE transactions on visualization and computer graphics}, 2019.

\bibitem{webster2014systematic}
David Webster and Ozkan Celik.
\newblock Systematic review of kinect applications in elderly care and stroke
  rehabilitation.
\newblock {\em Journal of neuroengineering and rehabilitation}, 11(1):108,
  2014.

\bibitem{wolf1985determining}
Alan Wolf, Jack~B Swift, Harry~L Swinney, and John~A Vastano.
\newblock Determining lyapunov exponents from a time series.
\newblock {\em Physica D: Nonlinear Phenomena}, 16(3):285--317, 1985.

\bibitem{xia2018nonlinear}
Guiyu Xia, Huaijiang Sun, Beijia Chen, Qingshan Liu, Lei Feng, Guoqing Zhang,
  and Renlong Hang.
\newblock Nonlinear low-rank matrix completion for human motion recovery.
\newblock {\em IEEE Transactions on Image Processing}, 27(6):3011--3024, 2018.

\bibitem{yan2018spatial}
Sijie Yan, Yuanjun Xiong, and Dahua Lin.
\newblock Spatial temporal graph convolutional networks for skeleton-based
  action recognition.
\newblock In {\em Thirty-Second AAAI Conference on Artificial Intelligence},
  2018.

\bibitem{yang2019spatio}
Jingyu Yang, Xin Guo, Kun Li, Meiyuan Wang, Yu-Kun Lai, and Feng Wu.
\newblock Spatio-temporal reconstruction for {3D} motion recovery.
\newblock {\em IEEE Transactions on Circuits and Systems for Video Technology},
  2019.

\bibitem{yoon2018gain}
Jinsung Yoon, James Jordon, and Mihaela Van Der~Schaar.
\newblock Gain: Missing data imputation using generative adversarial nets.
\newblock {\em International Conference on Machine Learning}, 2018.

\bibitem{zhang2018ista}
Jian Zhang and Bernard Ghanem.
\newblock {ISTA}-net: Interpretable optimization-inspired deep network for
  image compressive sensing.
\newblock In {\em Proceedings of the IEEE Conference on Computer Vision and
  Pattern Recognition}, pages 1828--1837, 2018.

\bibitem{zhang2019view}
Pengfei Zhang, Cuiling Lan, Junliang Xing, Wenjun Zeng, Jianru Xue, and Nanning
  Zheng.
\newblock View adaptive neural networks for high performance skeleton-based
  human action recognition.
\newblock {\em IEEE transactions on pattern analysis and machine intelligence},
  2019.

\end{thebibliography}
}

\appendix
\section{Network architecture and training details}

\subsection{HDM dataset \cite{cg-2007-2}}
\paragraph{Autoencoder training:} The encoder consists of 4 temporal convolution layers with filter size of 4, and the number of feature maps in each layer is set to $75$ (equal to the number of channels at the input layer). We use a latent space dimension of $200$. The decoder consists of 4 temporal transposed convolutional layers. The network parameters are trained using Adam optimizer \cite{kingma2014adam} for $2 \times 10^5$ iterations with a batch size of $64$ and an initial learning rate of $10^{-3}$. The learning rate is reduced by one-tenth after $1.5 \times 10^5$ and $1.8 \times 10^5$ iterations.

\paragraph{Classifier training:} We use a TCN classifier similar \cite{kim2017interpretable}. It consists of 3 TCN blocks with one convolutional layer each. The network is trained to minimize cross-entropy loss for $2 \times 10^5$ iterations with a batch size of $64$, and is optimized using Adam \cite{kingma2014adam} with an initial learning rate of $10^{-3}$ and reduced to one-tenth after $10^5$ and $1.8 \times 10^5$ iterations.

\subsection{NTU dataset \cite{shahroudy2016ntu}}
\paragraph{Autoencoder training:} 
The encoder consists of 3 temporal convolution layers with filter size of $8$, and the number of feature maps in each layer is set to $75$ (equal to the number of channels at the input layer). We experiment with latent space dimension $= 100,200$. The decoder consists of 3 temporal transposed convolutional layers. We use a latent space dimension of $200$. The decoder consists of 4 temporal transposed convolutional layers. The network parameters are trained using Adam optimizer \cite{kingma2014adam} for $2 \times 10^5$ iterations with a batch size of $64$ and an initial learning rate of $10^{-3}$. The learning rate is reduced by one-tenth after $1.5 \times 10^5$ and $1.8 \times 10^5$ iterations.

\paragraph{Classifier training:}
As the action classifier, we use a TCN classifier identical to that proposed by Kim and Reiter \cite{kim2017interpretable}. It consists of 3 TCN blocks with 3 convolutional layers each. The network is trained to minimize cross-entropy loss for $2 \times 10^5$ iterations with a batch size of $64$, and is optimized using Adam \cite{kingma2014adam} with an initial learning rate of $10^{-3}$ and reduced to one-tenth after $10^5$ and $1.8 \times 10^5$ iterations. 

\appendix
\section{Details about baselines and experiments}

\subsection{Sparse coding}
We use "MiniBatchDictionaryLearning" provided in the scikit-learn toolbox \cite{scikit-learn} which implements the algorithm by Mairal et al. \cite{mairal2009online} to first learn a dictionary based on the training set. We first create a matrix $D$ such that each row of the $D$ is a training action sequence where all the frames and joints are vectorized into a single vector. The dictionary is learned by performing essentially matrix factorization with additional constraints using an online learning approach: 

\begin{align}
    (U^*, V^*) &= \arg \min_{U, V} \frac{1}{2}\norm{D - UV}_2^2 + \alpha \norm{U}_1 \\ 
    \text{s.t.} & \quad \norm{V}_k = 1, \quad \forall 0 <= k <= n,
\end{align}

where $n$ is the number of components in the dictionary. We use $n = 500$ for both HDM and NTU datasets. Once the dictionary $V$ is learned, given a vectorized test action with unobserved joints, $X$, we find a sparse vector $\Theta = [\theta_1, \theta_2, \dots \theta_n]$ such that $X \approx \hat{X} = \sum_{i=1}^n \theta_i V_i$. $\hat{X}$ is used as the output of the action completion algorithm. $\Theta$ is computed using orthogonal matching pursuit.

\subsection{Frame-wise recovery}
The proposed method in this paper uses a temporal convolutional autoencoder that can exploit temporal correlations between frames in order to learn a more accurate representation of the action manifold. In order to test its importance, we consider a baseline where the deep prior only contains information about the manifold of human poses i.e., single frames, rather than action sequences. For this, we train an autoencoder for \emph{frames} with $8$ fully-connected layers with ReLUs  and the latent dimension is set to be equal to $8$. This model is very similar to that proposed by Holden et al. \cite{holden2018robust}. When all the frames of the action are considered together, the overall latent dimension, with the latent vectors concatenated, is of the same order as the latent dimension in the case of the temporal convolutional autoencoder for actions. These results are shown in Table 2 of the main paper as well as Tables \ref{table:hdm_ntu_results_3} and \ref{table:hdm_ntu_results_4} here.

\subsection{Different joints missing in different frames}
In the main paper, the focus was on recovering trajectories of joints/markers which are misssing completely throughout the action sequence. We focused on this setting because it is usually not studied in literature and it is more challenging as many of the temporal interpolation techniques become no longer applicable. However, the proposed method is readily applicable to the setting where different joints are missing in different frames. The entire algorithm is exactly the same, except the training phase where we randomly drop different joints from different frames in every action sequence. Table \ref{table:diff_joints_diff_frames} shows the results thus obtained for the HDM dataset at different train/test OTP ratios. As in the case of fully missing joints, the performance reduces gracefully when more joints are dropped, using the proposed latent space optimization approach. We also see that the reconstruction and classification results are nearly the same as when the joints are missing completely throughout the sequence (Table 1 in the main paper). 

\subsection{Additional baseline: denoising autoencoder}
In the main paper, in order to train autoencoders on training data that contain missing joints, we use the following Ambient AE loss function (Equation 2 in the main paper)
Instead, as an additional baseline, we train a denoising autoencoder. Here, we assume that we have access to a perfect training set with all joint trajectories intact. We train the autoencoder to map from sequences with missing joints to the ground-truth action sequences with information for all joints. Please note that this assumes that we have access to a clean training set. In the main paper, we work with a more challenging setting where even at training time, the ground-truth action sequences have some joints/markers missing. We provide comparisons between our proposed method (using the Ambient AE loss) and the denoising autoencoder for the HDM05 dataset in Table \ref{table:dae_results}. All the other architecture and training details are held constant. The results clearly show that the ambient AE framwork, which does not assume access to a clean training set, actually outperforms the denoising AE framwork. This is because, even though at training time the autoencoder is given complete ground-truth information, at test time, a completely different set of joints may be missing which was not seen at training time. This discrepancy between training and test time leads to poorer performance for the denoising AE. Also note that, for both denoising and ambient AE, latent space optimization at test time significantly improves the reconstruction both in terms of RMSE and classifier accuracy. 

\subsection{Self-similarity matrix}
\begin{figure*}[t]
\centering
\includegraphics[width=0.95\linewidth]{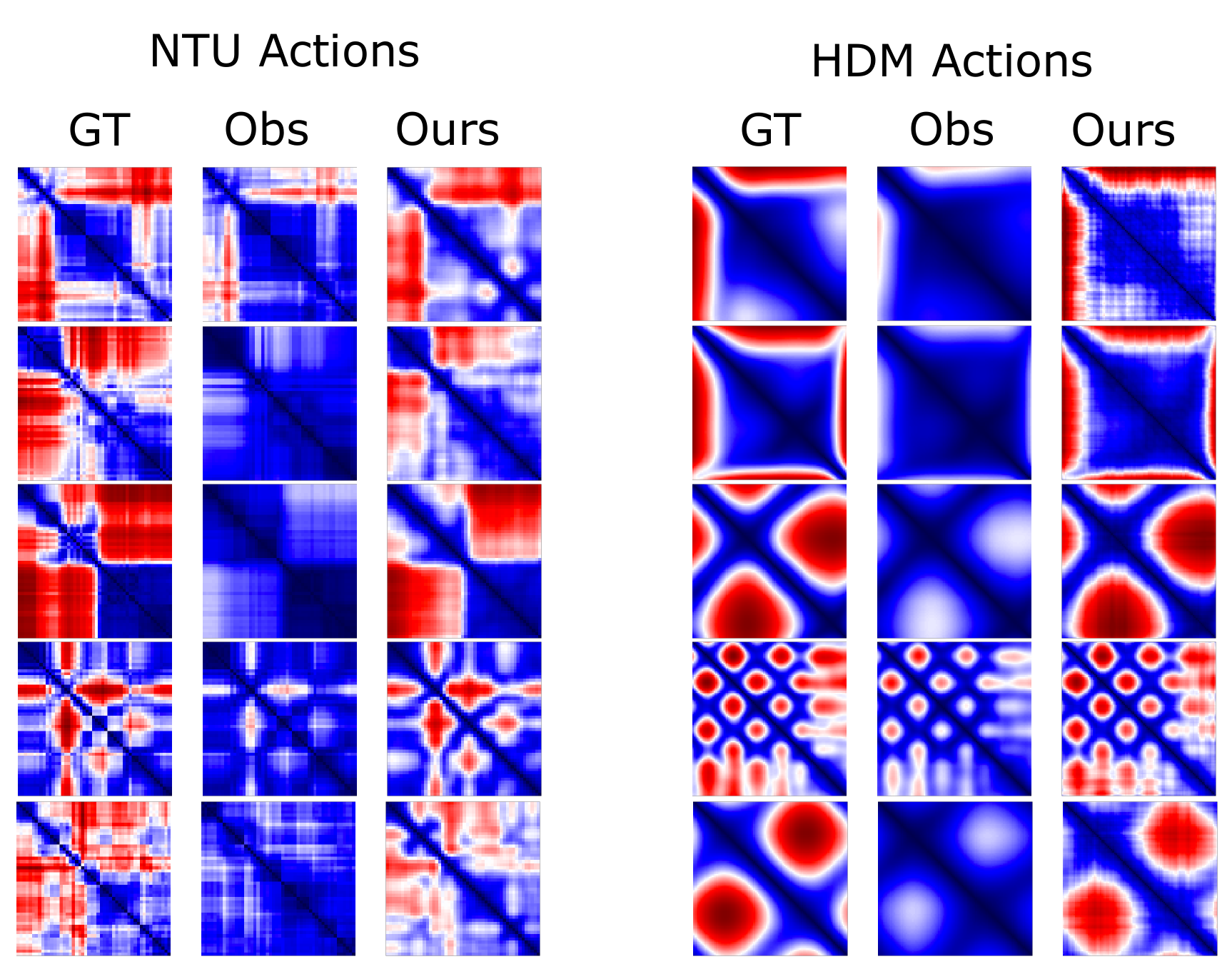}
\caption{\small{We visualize the dynamics for some actions using the self-similarity matrices (SSMs) on the two datasets for Train/Test OTP = 75/50. We see that even though a lot of dynamics are lost in the observed action with missing joints (Obs), compared to the ground-truth (GT) the proposed method (Ours) recovers them effectively. The images are normalized by the intensities in the ground-truth.}}
\label{fig:ntu_hdm_ssm}
\vspace{-0.25in}
\end{figure*}

In order to try and visualize the differences in the dynamics of the reconstructed actions for the baseline and proposed methods compared to the ground-truth, we use self-similarity matrices (SSMs) \cite{junejo2008cross}. SSMs capture dynamics better than using just classification accuracy, and at the same time, can be easily visualized. The SSM of a sequence X, $SSM(X) \in \mathbb{R}^{N \times N}$ is constructed using $SSM(X)_{n,m} = e^{-\frac{\norm{X_n - X_m}}{\sigma^2}}$, where $X_n$ is the $n^{th}$ frame of $X$ and $\sigma^2$ is the variance of the distances between all pairs of frames. Some visualizations of the SSMs for both HDM05 and NTU datasets can be seen in Figure \ref{fig:ntu_hdm_ssm}.

\section{Videos of recovered joint trajectories, additional results, t-SNE visualizations}

In this section, we show results and experiments performed that could not be added in the main paper due to space constraints. We have generated videos showing the reconstructions obtained using the proposed latent space optimization approach as compared to a single feedforward pass through the autoencoder. We show results for both HDM and NTU datasets at two different Train/Test OTPs which show that the proposed method outperforms the baseline and produces recovers accurate motion sequences for the unobserved joints:

\begin{enumerate}
    \item Train/Test OTP = 75/50: HDM\_results\_Train\_OTP\_75\_Test\_OTP\_50.avi, \\  NTU\_results\_Train\_OTP\_75\_Test\_OTP\_50.avi
    \item Train/Test OTP = 50/50: HDM\_results\_Train\_OTP\_50\_Test\_OTP\_50.avi, \\    NTU\_results\_Train\_OTP\_50\_Test\_OTP\_50.avi
\end{enumerate}

\begin{figure*}[!htb]
\centering
\subfloat[Train/Test OTP = 100/50]{\includegraphics[width=\textwidth]{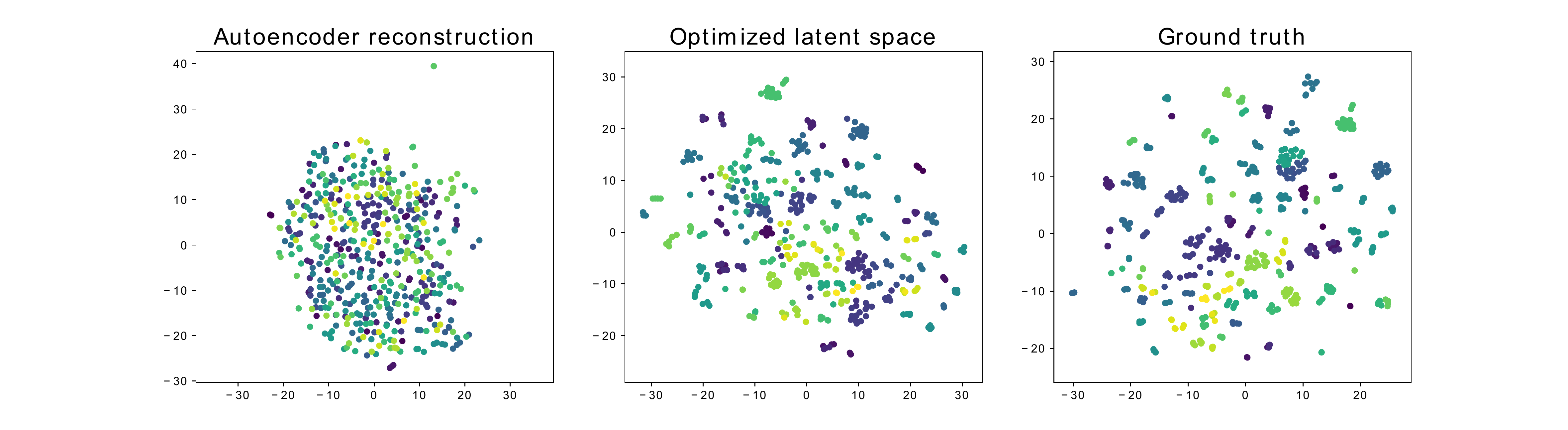}\label{fig:hdm_tsne_1}}

\subfloat[Train/Test OTP = 75/50]{\includegraphics[width=\textwidth]{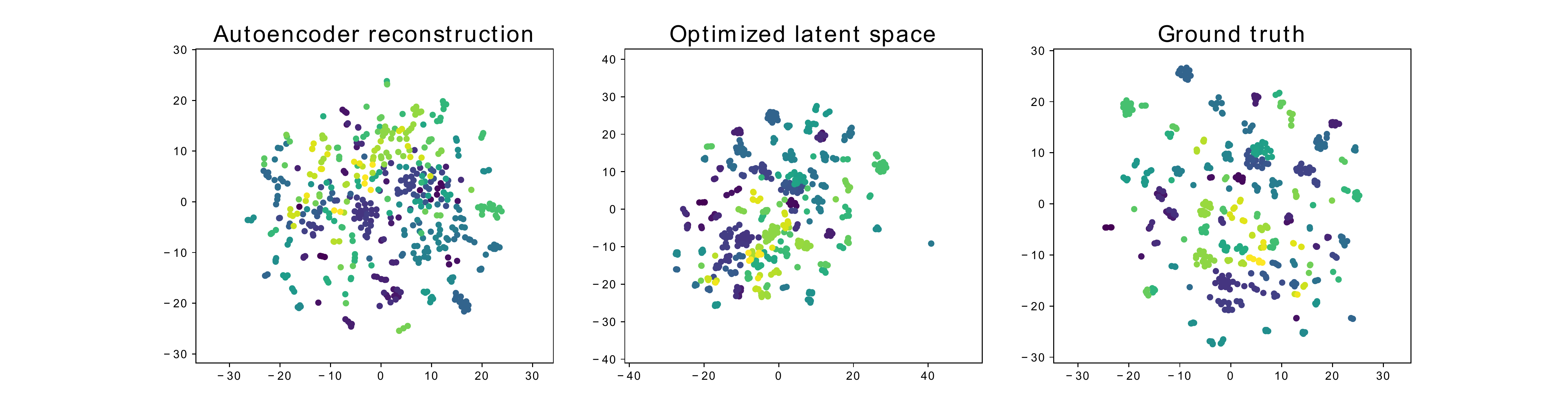}\label{fig:hdm_tsne_2}}

\subfloat[Train/Test OTP = 50/50]{\includegraphics[width=\textwidth]{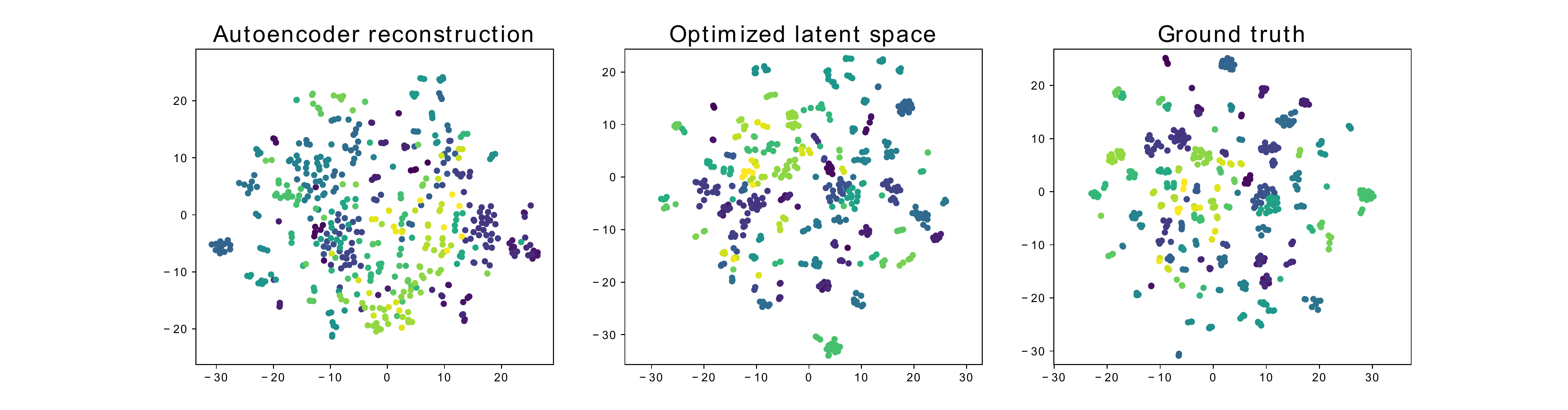}\label{fig:hdm_tsne_3}}
\caption{\small{t-SNE embeddings of the penultimate layer features of the action classifier for Fold 1 HDM test set for different Train/Test OTPs. We see that a lot of semantic information is lost when the joints are dropped, but can be recovered most effectively with an optimized latent space. Note that different runs of the t-SNE algorithm can produce slightly different results, however the overall trend remains the same.}}
\label{fig:hdm_results_tsne}
\end{figure*}

\begin{figure*}[!htb]
\centering
\subfloat[Train/Test OTP = 100/50]{\includegraphics[width=\textwidth]{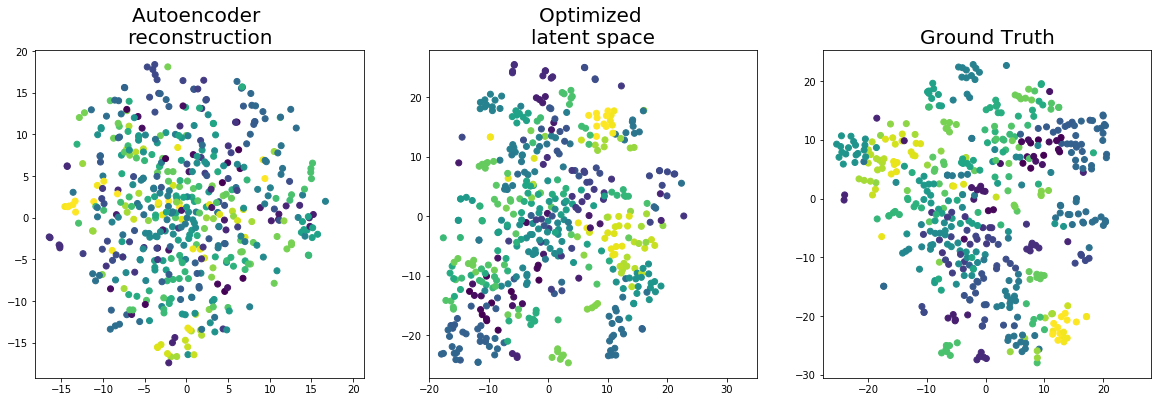}\label{fig:ntu_tsne_1}}

\subfloat[Train/Test OTP = 75/50]{\includegraphics[width=\textwidth]{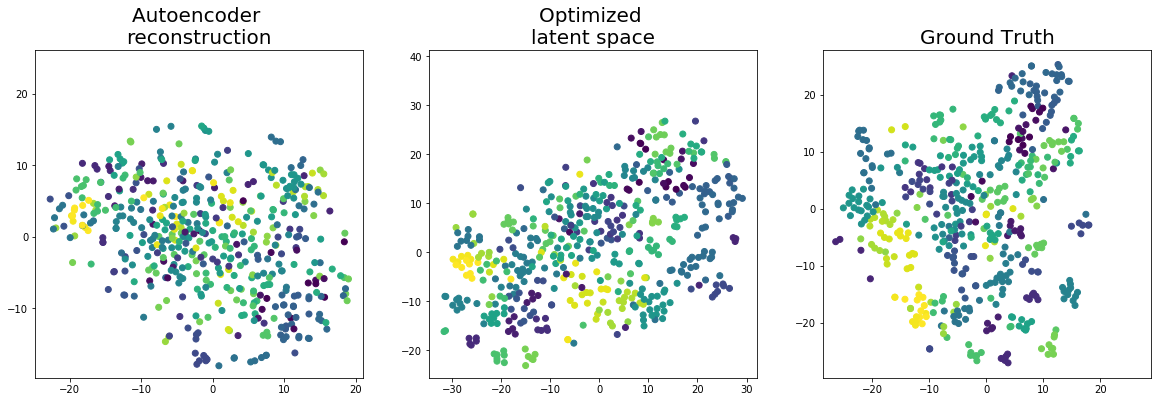}\label{fig:ntu_tsne_2}}

\subfloat[Train/Test OTP = 50/50]{\includegraphics[width=\textwidth]{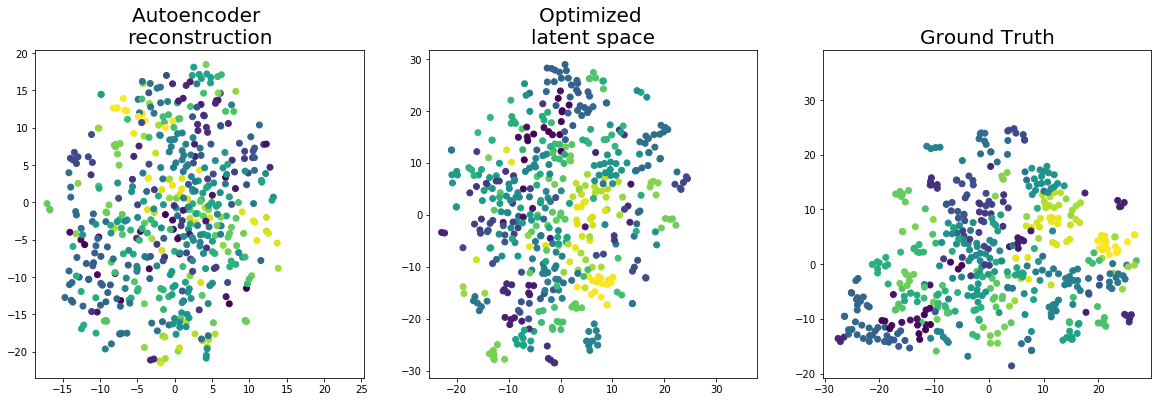}\label{fig:ntu_tsne_3}}
\caption{\small{t-SNE embeddings of the penultimate layer features of the action classifier for 500 randomly sampled actions from the NTU test set for different Train/Test OTPs.We see that a lot of semantic information is lost when the joints are dropped, but can be recovered most effectively with an optimized latent space.}}
\label{fig:ntu_results_tsne}
\end{figure*}

Table 2 in the main paper shows the comparison of the proposed method with multiple baselines for Train/Test OTP = 75/50. Here, in Tables \ref{table:hdm_ntu_results_3} and \ref{table:hdm_ntu_results_4}, we show similar comparison for 100/50 as well as 50/50 respectively. 

t-SNE \cite{maaten2008visualizing} visualizations for the penultimate layer features of the action classifier network show that the proposed method of latent space optmiziation outperforms the baselines in terms of producing more class-discriminative clusters. The visualizations for Fold 1 of the HDM test set at different Train/Test OTPs are shown in Figure \ref{fig:hdm_results_tsne}. The visualizations for the NTU test set at different Train/Test OTPs are shown in Figure \ref{fig:ntu_results_tsne}.


\begin{table*}[t]
\small
\centering
\begin{tabular}{c|c|c|c|c}
Method & \multicolumn{2}{c|}{HDM} & \multicolumn{2}{c}{NTU} \\
\cline{2-5}
& RMSE (cm) & Acc (\%) & RMSE (cm) & Acc (\%) \\
\hline
Sparse Coding & 17.79 & 11.89 & 18.23 & 8.19 \\
\hline
Frame-wise $D_F(E_F(Y_F))$ & 21.41 & 11.10 & 20.22 & 8.80\\
\hline
Frame-wise $D_F(\mathbf{z_F}^*)$ & 21.81 & 13.07 & 20.25 & 9.76 \\
\hline
Action $D(E(Y))$ & 9.87 & 28.99 & 9.39 &  31.89\\
\hline
\makecell{Action $D(\mathbf{z}^*)$ \\ (Proposed)} & \textbf{2.99} & \textbf{73.47} & \textbf{5.19} & \textbf{65.15} \\
\end{tabular}
\caption{Experimental results for HDM05 (averaged over 5 folds) and NTU datasets compared to different baselines for train/test OTP = 100/50. We observe easily that the proposed optimization-based reconstruction is superior to all the baselines considered. $D_F$ and $E_F$ refer to the fact that the encoder and decoder operate on a single frame at a time, rather than an action sequence.}
\label{table:hdm_ntu_results_3}
\end{table*}


\begin{table*}[t]
\small
\centering
\begin{tabular}{c|c|c|c|c}
Method & \multicolumn{2}{c|}{HDM} & \multicolumn{2}{c}{NTU} \\
\cline{2-5}
& RMSE (cm) & Acc (\%) & RMSE (cm) & Acc (\%) \\
\hline
Sparse Coding & 20.35 & 10.42 & 19.37 & 7.01\\
\hline
Frame-wise $D_F(E_F(Y_F))$ & 21.77 & 11.30 & 20.57 & 8.41 \\
\hline
Frame-wise $D_F(\mathbf{z_F}^*)$ & 21.28 & 13.14 & 20.21 & 9.64 \\
\hline
Action $D(E(Y))$ & 8.77 & 43.56 & 9.37 & 34.94 \\
\hline
\makecell{Action $D(\mathbf{z}^*)$ \\ (Proposed)} & \textbf{3.05} & \textbf{74.37} & \textbf{5.29} & \textbf{64.59} \\
\end{tabular}
\caption{Experimental results for HDM05 (averaged over 5 folds) and NTU datasets compared to different baselines for train/test OTP = 50/50. We observe easily that the proposed optimization-based reconstruction is superior to all the baselines considered. $D_F$ and $E_F$ refer to the fact that the encoder and decoder operate on a single frame at a time, rather than an action sequence.}
\label{table:hdm_ntu_results_4}
\end{table*}


\begin{table*}[htb]
\small
\centering
\begin{tabular}{c|c|c|c|c|c}
Train OTP / Test OTP & Method & \multicolumn{2}{c|}{Denoising AE loss} & \multicolumn{2}{c}{Ambient AE Loss (proposed)} \\
\cline{3-6}
& & RMSE (cm) & Acc (\%) & RMSE (cm) & Acc (\%)\\
\hline

\multirow{2}{*}{75/75} & $D(E(Y))$ & 9.13 & 47.69 & 6.07 & 68.72 \\
& $D(\mathbf{z}^*)$ & 5.21 & 71.67 & \textbf{2.27} & \textbf{78.61} \\
\hline

\multirow{2}{*}{75/50} & $D(E(Y))$ & 11.76 & 22.98 & 8.52 & 44.81\\
& $D(\mathbf{z}^*)$ & 7.66 & 55.98 & \textbf{2.98} & \textbf{74.71}\\
\hline

\multirow{2}{*}{50/50} & $D(E(Y))$ & 12.70 & 19.29 & 8.77 & 43.55\\
& $D(\mathbf{z}^*)$ & 9.98 & 46.42 & \textbf{3.05} & \textbf{74.37} \\
\end{tabular}
\caption{\small{Comparison between denoising AE loss v/s ambient AE loss, for HDM05 at different train/test OTPs in terms of RMSE and action recognition accuracy (Acc). We observe easily that the proposed method is far superior to the denoising AE loss function. Latent space optimization proposed in the paper is useful in both cases. Note that when Train OTP = 100, both loss functions are identical and thus yield the same results}}
\vspace{-0.2 in}
\label{table:dae_results}
\end{table*}


\begin{table*}[htb]
\small
\centering
\begin{tabular}{c|c|c|c}
Train OTP / Test OTP & Method & RMSE (cm) & Acc (\%) \\
\hline

100/100 & $D(E(Y))$ & 3.48 & 79.23 \\
\hline

\multirow{2}{*}{100/75} & $D(E(Y))$ & 6.27 & 60.60 \\
& $D(\mathbf{z}^*)$ & \textbf{2.16} & \textbf{78.06}\\
\hline

\multirow{2}{*}{100/50} & $D(E(Y))$ & 10.03 & 27.99 \\
& $D(\mathbf{z}^*)$ & \textbf{2.99} & \textbf{73.65}\\
\hline

\multirow{2}{*}{75/75} & $D(E(Y))$ & 5.46 & 69.27\\
& $D(\mathbf{z}^*)$ & \textbf{2.21} & \textbf{78.46} \\
\hline

\multirow{2}{*}{75/50} & $D(E(Y))$ & 8.11 & 43.31 \\
& $D(\mathbf{z}^*)$ & \textbf{3.02} & \textbf{73.63} \\
\hline

\multirow{2}{*}{50/50} & $D(E(Y))$ & 7.37 & 56.55 \\
& $D(\mathbf{z}^*)$ & \textbf{3.10} & \textbf{73.40} \\
\end{tabular}
\caption{\small{Experimental results for HDM05 (averaged over 5 folds) when different joints are missing for different frames, for varying train/test OTPs in terms of RMSE and action recognition accuracy (Acc). We observe easily that the proposed optimization-based reconstruction is far superior to a feedforward pass through the autoencoder. As the train OTP is reduced, performance degrades more gracefully in the case of the optimization-based approach. In all cases, we can get to within $5 \%$ points of the oracle action recognition performance (train /test OTP = $100/100$).}}
\vspace{-0.15in}
\label{table:diff_joints_diff_frames}
\end{table*}


\end{document}